\documentclass{article}

     \PassOptionsToPackage{numbers, compress}{natbib}

\usepackage[preprint]{neurips_2022}

\usepackage[utf8]{inputenc} 
\usepackage[T1]{fontenc}    
\usepackage{hyperref}       
\usepackage{url}            
\usepackage{booktabs}       
\usepackage{amsfonts}       
\usepackage{dsfont}
\usepackage{nicefrac}       
\usepackage{microtype}      
\usepackage{xcolor}         
\usepackage{amsmath,bm}
\usepackage{upgreek}
\usepackage{tikz}
\usepackage{booktabs}
\usepackage{nicefrac} 
\usepackage{pgfplots}
\usepackage{microtype}      
\title{Preserving Fine-Grain Feature Information in Classification via Entropic Regularization}
\usepackage{caption}
\usepackage{subfigure}
\usepackage{graphicx}
\usepackage{enumitem}
\author{Raphael Baena \qquad Lucas Drumetz \qquad Vincent Gripon \\ 
IMT Atlantique, Lab-STICC, UMR CNRS 6285, F-29238, France\\
\texttt{surname.lastname@imt-atlantique.fr}
}

\begin{document}
\maketitle

\begin{abstract}
Labeling a classification dataset implies to define classes and associated coarse labels, that may approximate a smoother and more complicated ground truth. For example, natural images may contain multiple objects, only one of which is labeled in many vision datasets, or classes may result from the discretization of a regression problem. Using cross-entropy to train classification models on such coarse labels is likely to roughly cut through the feature space, potentially disregarding the most meaningful such features, in particular losing information on the underlying fine-grain task. In this paper we are interested in the problem of solving fine-grain classification or regression, using a model trained on coarse-grain labels only.  We show that standard cross-entropy can lead to overfitting to coarse-related features. We introduce an entropy-based regularization to promote more diversity in the feature space of trained models, and empirically demonstrate the efficacy of this methodology to reach better performance on the fine-grain problems. Our results are supported through theoretical developments and empirical validation.

\end{abstract}

\section{Introduction}

Solving regression or fine-grain classification problems can be performed in two ways: learning to predict the actual fine-grain value associated with an input, that we call a fine label in the following, or transforming the fine labels into coarse labels creating an ad-hoc classification problem that can be used as a proxy to the fine-grain problem~\cite{ageintoregression,ageintoregressionandclassif}. Using this proxy classification problem can be useful if it is easier to solve than the original problem, or if fine grain ground truth labels are hard or expensive to obtain. Reciprocally, in many contexts classification problems could be seen as a discretization of an underlying more subtle regression problem where labels are not necessarily finite~\cite{coarselabels}. Having access to these refined labels could improve generalization performance, as is illustrated with the use of distillation techniques~\cite{distilling}.

In this paper, we are interested in showing it is possible to partially uncover the disregarded fine labels when training a model on the derived coarse labels, showing an ability of machine learning techniques to generalize beyond a considered classification task and to solve even more subtle problems. Such ability is not granted for free. Indeed, as a model is trained more and more specifically to predict the coarse labels, it is expected that it loses such an ability of finer generalization, as a result of overfitting to the training task. This phenomenon is often observed in the case of transfer learning, in particular with few-shot problems, where early stopping of the training can lead to better performance on subsequent tasks. This can be explained by the two-phase behavior of the cross-entropy loss training described in~\cite{blackbox}, where mutual information between the feature space and the inputs is lost in the second phase as the focus is shifted towards excelling at the training task.  

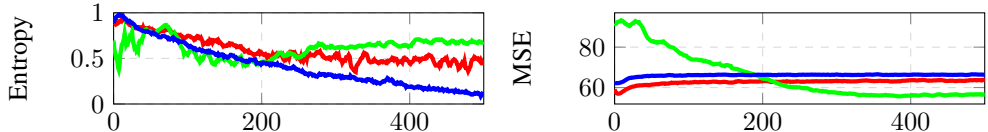
\begin{figure}
\centering
  \subfigure{
  \begin{tikzpicture}
  \pgfplotsset{%
    width=.1pt,
    height=.2\textwidth
}
      \begin{axis}[
          width=6.5cm, 
          grid=major, 
          grid style={dashed,gray!30}, 
          ylabel= Entropy ,
          ytick={0,0.5,1},
          legend style={at={(0.5,-0.2)},anchor=north}, 
           ymin=0,
           ymax=1,
           xmax=500,
           xmin=0,
        ]
        \addplot [  red,   ultra thick] table[x=epochs,y=CE,col sep=comma] {evolution_entropy.csv}; 
        \addplot [  green,   ultra thick] table[x=epochs,y=FIERCE,col sep=comma] {evolution_entropy.csv}; 
        \addplot [  blue,   ultra thick] table[x=epochs,y=LS,col sep=comma] {evolution_entropy.csv}; 
      \end{axis}
    \end{tikzpicture}
}
  \subfigure{
  \begin{tikzpicture}
  \pgfplotsset{%
    width=.1pt,
    height=.2\textwidth
}
      \begin{axis}[
          width=6.5cm, 
          grid=major, 
          grid style={dashed,gray!30}, 
          ylabel= MSE,
          legend style={at={(0.5,-0.2)},anchor=north}, 
        xmax=500,
        xmin=0,
        ]
        \addplot [  red,   ultra thick] table[x=epochs,y=CE,col sep=comma] {evolution_mse.csv}; 
        \addplot [ green,   ultra thick] table[x=epochs,y=FIERCE,col sep=comma] {evolution_mse.csv}; 
        \addplot [ blue,   ultra thick] table[x=epochs,y=LS,col sep=comma] {evolution_mse.csv}; 
      \end{axis}
    \end{tikzpicture}
}
    \caption{Evolution of the Entropy of the feature space (rescaled between 0 and 1) and Mean Square Error on the age regression dataset~\cite{ageestimation} for the different criteria: Cross Entropy (red), FIERCE (proposed) with $\lambda = 0.3$ (green), Label Smoothing (blue). We observe that models trained with standard cross-entropy or label smoothing briefly display a minimal MSE before stabilizing to a larger value. On the contrary, the proposed FIERCE method reaches an overall lower MSE that is maintained through additional training epochs. This ability to reach and maintain lower MSE is negatively correlated with the entropy measured in the feature space, shown in the left figure.}
  \label{fig:EvolutionEntropyMSE}
\end{figure}

We observed a very similar behavior when it comes to uncover the fine labels from a training on coarse labels. We ran a simple experiment on age estimation~\cite{ageestimation} with a ResNet~\cite{resnet}. We are given coarse ranges of ages in the training set as the training labels while being interested in predicting the precise age of an individual. We defined only two classes for the training: more or less than 36 years old. At each epoch we monitored the entropy of the feature space and in parallel we looked at the ability of an optimal transport methodology to retrieve the exact age (fine labels) from one-dimensional feature vectors. We depict the two measures in Figure~\ref{fig:EvolutionEntropyMSE}. In a first phase, the Mean Square Error (MSE) of the regression task decreases with the classification error rate. During a second phase, much longer, the MSE quickly reaches a minimum then significantly increases while the classification error rate remains stable. By monitoring the evolution of the entropy of the feature space, it appeared clearly that the MSE and the Entropy of the feature space were strongly correlated. 

Therefore, as an aim to improve generalization to subtle unseen labels, we introduce in this paper Feature Information Entropy Regularized Cross Entropy (FIERCE): an entropy-based regularization defined on the feature space of trained architectures. This criterion is meant to ensure that this once gained ability is not compromised with further training. As a result, trained models show better generalization capabilities out of the scope of the training task, as we demonstrate with multiple experiments. Interestingly, it is not required to rely on early-stopping anymore as shown on Figure~\ref{fig:EvolutionEntropyMSE}.

The main motivating problem of this paper is that of having only access to coarse labels while being interested in a more subtle discrimination of the data. Such scenarios are likely to occur in practical applications, as coarse labeling can be significantly less expensive, or even simple freely available~\cite{fewlabel}. It is also related to the problem of transfer learning, where features are used to process a new task. On the problem of age regression~\cite{ageintoregression}, we show that even when we only distinguish two ranges of ages (more or less than 36 years old), it is possible to unveil a more precise estimation of the real age of an individual. We also demonstrate the ability of the proposed method to improve performance in classical transfer problems, including few-shot settings.

The main claims of our work are as follows:
\begin{itemize} [topsep=-3pt,itemsep = -0.17cm,noitemsep]

    \item We introduce the problem of coarse-to-fine classification, which consists in learning from coarse labels to generalize to finer --unknown during training-- labels.
    \item We experimentally show and theoretically motivate the fact that entropy in the feature space and performance on the fine labels task are strongly bound to each other.
    \item We introduce a regularization method that promotes a high entropy for the features, relying on the softmax-gumbel function and a fixed random embedding in the feature space.
    \item Within a Bayesian Framework we derive the gradients of Cross Entropy, Label Smoothing and our method to understand their impact on the feature space during training.
    \item We demonstrate the ability of the proposed method to partially unveil fine labels using multiple datasets, and showcase transfer learning potential.
\end{itemize}

We provide free access to our codes to be used to reproduce results and/or propose new developments.




\section{Related Work}
\subsection{The Cross-Entropy in classification problem}
A deep neural network (DNN) can be represented by a function $f_{\boldsymbol{\theta}}:\mathbf{x} \mapsto \mathbf{y}$, here $f_{\boldsymbol{\theta}}$ outputs a soft decision associated with its input $\mathbf{x}$. It is usually obtained by composing simple functions called layers~\cite{goodfellow}, resulting in possibly very complex architectures. In the context of supervised learning, it is typical to name the output space of the penultimate layer as the \emph{feature} space. The prediction of the DNN is usually carried out by a linear classifier applied on the features $r_{\boldsymbol{\theta}}(\mathbf{x})$ of the input $\mathbf{x}$. A consequence of this paradigm is that we are interested in a DNN capable of providing features from which we can retrieve accurate prediction of the labels $\mathbf{y}$. 

DNN parameters $\boldsymbol{\theta}$ are tuned using a training set $\mathcal{D}_{train}$ with the intent of generalizing well to unseen data. The training of the network relies on the minimization of a criterion function $\mathcal{L}$ where the parameters of the DNN are updated proportionally to the gradient of the loss with the goal of minimizing the following criterion:
\begin{align}
    \underset{\boldsymbol{\theta}}{\textrm{min}} \ \mathbb{E}_{\mathcal{D}_{train}}  [ \mathcal{L}(\mathbf{x},\mathbf{y},\boldsymbol{\theta})], \text{ where } \mathcal{D}_{train} \text{ is the training set.}
\end{align}
In classification, the DNN's output $f_{\boldsymbol{\theta}}(\mathbf{x})$ can be interpreted as the conditional class probabilities $q_{\boldsymbol{\theta}}(\mathbf{y}_i |\mathbf{x})$, after a softmax function has been applied~\cite{distilling}:
\begin{align}
    q_{\boldsymbol{\theta}}(y_i |\mathbf{x}) = \frac{\exp(f_{\boldsymbol{\theta}}(\mathbf{x})_i/ \tau )}{\sum_j \exp(f_{\boldsymbol{\theta}}(\mathbf{x})_j/ \tau)}.
\end{align}
Thus $q_{\boldsymbol{\theta}}(\mathbf{y} |\mathbf{x})$ is the inferred distribution that hopefully fits well the real distribution $p_{\boldsymbol{\theta}}(\mathbf{y} |\mathbf{x})$.

The choice of the criterion has a direct impact on the training and the final characteristics of the neural network~\cite{lossimpact}. In particular, it can affect the features $r_{\boldsymbol{\theta}}(\mathbf{x})$~\cite{whenlabelsmoothinghelp,islabelsmoothingtrully} and the output distributions $q_{\boldsymbol{\theta}}(\mathbf{y} |\mathbf{x})$~\cite{penalisingconfident,distilling,calibration,entropylabelsmoothing}. Among the different criteria, the Cross Entropy loss is widely used in supervised learning when dealing with classification problems~\cite{goodfellow}. Actually, maximizing the likelihood of a categorical distribution of class labels whose probabilities are given by the $q_{\boldsymbol{\theta}}(y_i |\mathbf{x})$  is equivalent to  minimizing the cross-entropy loss~\cite{bishop}:
\begin{align}\label{eq:crossentropy}
     \mathcal{L}_{CE}(\mathbf{x},\mathbf{y},\boldsymbol{\theta}) =  \mathbb{E}_{\mathbf{x}} \left[ \sum _i y_i \log( q_{\boldsymbol{\theta}}(y_i |\mathbf{x})) \right], \text{where } y_i = \left\{ \begin{array}{ll}
                  1 \text{ if } i  \text{ is the class of } \mathbf{x}\\
                  0 \text{ otherwise}\\
                \end{array}\right. .
\end{align}

Minimizing the Cross Entropy with the Stochastic Gradient Decent (SGD) optimization algorithm produces two main sequential phases~\cite{blackbox}. During the first phase, the mutual information $I(\mathbf{y},\mathbf{r})$ between the features $\mathbf{r}$ and the output $\mathbf{y}$ increases. During the second phase, usually much longer, the same mutual information $I(\mathbf{x},\mathbf{r})$ between the input and the features decreases. In other words, the neural network first finds the features on which it can predict the labels before compressing them in order to keep only the most relevant pieces of information for the classification task. 


In~\cite{calibration}, the authors showed that the probabilistic error and miscalibration worsen as classification error is reduced by the DNN. This miscalibration could be an indicator of the lost information in the feature space. A typical way to reduce miscalibration consists in adding a regularization $\mathcal{R}(\mathbf{x},\mathbf{y},\boldsymbol{\theta})$ to the loss function \cite{calibration}, resulting in the following optimization problem:
\begin{align}
    \underset{\theta}{\textrm{min}} \ \mathbb{E}_{\mathcal{D}_{train}}  [ \mathcal{L}_{CE}(\mathbf{x},\mathbf{y},\boldsymbol{\theta})) +\mathcal{R}(\mathbf{x},\mathbf{y},\boldsymbol{\theta}) ].
\end{align}
Regularization involves a large range of techniques acting on weights of the DNNs, for example batch normalization~\cite{batchnorm}, weight decay~\cite{weightdecay} and dropout~\cite{dropout}. 

\subsection{Entropy and output regularization}
Some regularization techniques can be applied directly on the output distribution as a criterion: e.g confidence penalization~\cite{penalisingconfident}, Label Smoothing~\cite{labelsmoothing} or distillation~\cite{distilling}. Multiple authors showed that these techniques actually rely on an entropy regularization of the output distribution~\cite{penalisingconfident,entropyregularizationfewdatapeakness,entropylabelsmoothing}. In the next paragraphs, we give more details about these techniques. 

\textbf{Confidence penalization}~\cite{penalisingconfident} consists in adding a regularization term made of the negative entropy of the output distribution $ \mathcal{R}(\mathbf{x},\mathbf{y},\boldsymbol{\theta}) = - H(q_{\boldsymbol{\theta}}(\mathbf{y} |\mathbf{x})) = $. The idea is to penalize confident output distributions as they typically correspond to low entropy distributions. As a result the inferred output distribution is smoothed, leading to increased generalization performance in some conditions.

\textbf{Label Smoothing}~\cite{labelsmoothing} is another technique that has experimentally been demonstrated to produce similar effects to confidence penalization~\cite{penalisingconfident}. Label Smoothing is a form of output regularization which smooths the target distribution $p(\mathbf{x}|\mathbf{y})$. In classification, this distribution is represented by coarse labels encoded by one-hot vectors, where the correct label probability is 1 and all other labels have probability 0. In the case of uniform Label Smoothing with coefficient $\sigma$, a boost in probability of $1-\sigma$ is assigned to the correct label and a penalty of $\sigma/(c-1)$ to the probabilities of other labels ($c$ is the number of classes). In~\cite{entropylabelsmoothing}, authors proved that Label Smoothing is a form of output entropy regularization, similarly to confidence penalization. More precisely, they showed that Label Smoothing is equivalent to adding a weighted Kullback-Leibler (KL) divergence between the uniform distribution $\mathcal{U}$ and $q_{\boldsymbol{\theta}}(y |\mathbf{x})$: $R(\mathbf{x},\mathbf{y},\boldsymbol{\theta}) = \mathcal{D}_{KL} ( \mathcal{U} || q_{\boldsymbol{\theta}}(\mathbf{y} |\mathbf{x}) )$. 

\textbf{Distillation} consists in using the class probabilities given by a (typically large) model to train another (typically smaller) model. These probabilities are considered as refined labels. When these labels have high entropy they are supposed to provide more information than the coarse labels \cite{distilling}. Distillation can be interpreted as a version of Label Smoothing where the smoothing is not uniform \cite{distiandlabelsmoothing,selfdisti}. The output distribution of the teacher is interpreted as a prior on the fine target distribution. In uniform Label Smoothing there is no information on the fine labels distribution: the prior is uniform \cite{selfdisti}. The authors of \cite{entropyregularizationfewdatapeakness,Dubey} argued that distillation enables neural networks to learn more generalizable features and encourage the classifier to reduce the specificity of the features. Differently to Distillation, other methods propose also to infer the smooth labels, e.g. in \cite{smoothlabelrec}.

Entropy regularized outputs are strong regularizers in supervised learning; these outputs can lead to smoother and more accurate output distributions~\cite{penalisingconfident,entropyregularizationfewdatapeakness}. These methods have also an impact of the geometry of the feature space~\cite{islabelsmoothingtrully,whenlabelsmoothinghelp}. The use of Label Smoothing encourages samples of the same class to lie in tight clusters and erases the information between the intra-class logits. In distillation the use of a teacher trained with Label Smoothing raised a controversy as the loss of information may be detrimental \cite{whenlabelsmoothinghelp}. One the other side, the authors of \cite{islabelsmoothingtrully} claimed that Label Smoothing allows to distinguish classes which are semantically close, which can be beneficial in the context of classification. More recently, the authors in \cite{betterloss} showed that ``better losses'' such as Label Smoothing, with greater class separation, are detrimental to the transfer performance as they produce less generic features. Similarly, we show in our experiments that Label Smoothing is actually detrimental if one wants to reuse the inferred features to perform regression tasks. 

Contrary to these works, our rationale is to promote entropy in the feature space instead of the output space, which we motivate in the following section.

\section{Impact of the criteria over the feature space }

Using a Bayesian framework, let us illustrate why promoting entropy in the feature space could lead to better performance in underlying fine labels tasks.

\subsection{Links between feature selection and Cross-Entropy and Label Smoothing}
In classification settings, a neural network typically first transforms inputs into feature vectors (hopefully linearly separable). These features are then classified with a logistic regression. We note $y$ the class of a sample, i.e, $y = \text{argmax}_i \mathbf{y}[i]$.
We consider a Bayesian framework to simplify our equations: the classification is given by $p ( y = y_i | \mathbf{r})$ where $\mathbf{r}$ are the features w.r.t to $\mathbf{x}$: $p_{\boldsymbol{\theta}}(\mathbf{r} |\mathbf{x})$ inferred by the network.  For simplicity, we take the classification layer as fixed in the equation; note that this not change anything to the generality of the proof since the gradient is updated using the chain-rule. With this framework the model first samples features $\mathbf{r}$ according to $\mathbf{x}$: $p_{\boldsymbol{\theta}}(\mathbf{r} |\mathbf{x})$; then it attributes  the class probabilities $\log ( p ( y = y_i | \mathbf{r}))$ corresponding to the output probability given by the classifier. In the following equations, we assume that the derivatives of the distributions are bounded by integrable functions, to be able to permute integrals and derivatives. \\

\textbf{Cross Entropy:} Within the Bayesian framework, one can rewrite Eq~\ref{eq:crossentropy} as:
\begin{align}
   \mathcal{L}^{Bayesian}_{CE}(\mathbf{x},\mathbf{y},\boldsymbol{\theta}) =  \mathbb{E}_\mathcal{D} \left[ - \mathbb{E}_{\mathbf{r} \sim p_{\boldsymbol{\theta}}(\mathbf{r} |\mathbf{x})} [ \log ( p ( y = y_i | \mathbf{r})) ] \right], \text{where } y_i \text{ is the coarse label of }\mathbf{x}.
\end{align}

Now computing the gradient of $\mathcal{L}^{Bayesian}_{CE}(\mathbf{x},\mathbf{y},\boldsymbol{\theta})$ w.r.t to $\boldsymbol{\theta}$, one gets:
\begin{align}
   \nabla_{\boldsymbol{\theta}}  \mathcal{L}^{Bayesian}_{CE}(\mathbf{x},\mathbf{y},\boldsymbol{\theta}) =  \mathbb{E}_\mathcal{D} \left[ - \int \nabla_{\boldsymbol{\theta}} (p_{\boldsymbol{\theta}}( \mathbf{r} |\mathbf{x})) \log ( p ( y = y_i | \mathbf{r}) d\mathbf{r}   \right].
\end{align}
Interestingly we note that $ \nabla_{\boldsymbol{\theta}} (p_{\boldsymbol{\theta}}( \mathbf{r} |\mathbf{x}))$ is weighted by $\log ( p ( y = y_i |\mathbf{r}) )$ meaning that the more the features $\mathbf{r}$ are selective for the classifier, the more they will be encouraged to be sampled. 

One way to visualize this effect on the gradient is to imagine a classification problem with 2 classes and a fixed classifier. Consider that the neural network can infer a 1d feature space where features are linearly separable by the classifier. The further the features $\mathbf{r}$ are from the decision boundary, the more important $\log ( p ( y = y_i | \mathbf{r})) $ is. So despite having  already linearly separable features, the gradient of the Cross Entropy will eventually force the features to be as far as possible to the margin and collapse into the two extremes of the feature space. Eventually the cross-entropy will cause overconfidence in the network prediction, which is well reported by \cite{calibration}. \\

 \textbf{Label Smoothing:} A similar derivation can be conducted for Label smoothing with coarse labels $\mathbf{y}$ uniformly smoothed by $\sigma< 0.5$:
 $\mathbf{y} = (1-\sigma) \mathbf{y} + \sigma \mathbf{1}$, or $   y[i] =\left\{ 
                \begin{array}{ll}
                   (1-\sigma) \text{ if } i   \text{ is the class of x},\\
                 \sigma \text{ otherwise}\\
                \end{array}
              \right.$, where $\mathbf{1} = [1, 1, \cdots, 1]^T$. The gradient becomes:
\begin{align}
    \mathbb{E}_{\mathbf{x}} \left[ - \int \nabla_{\boldsymbol{\theta}} (p_{\boldsymbol{\theta}}( \mathbf{r} |\mathbf{x})) \left[ (1-\sigma) \log ( p ( y = y_i | \mathbf{r}) + \frac{\sigma}{c-1} \log( \prod_{j, j\neq i} p ( y = y_j | \mathbf{r}))\right] d\mathbf{r}   \right].
\end{align}

One can note that $\log ( p ( y = y_i | \mathbf{r}))$ is reduced by a factor $(1-\sigma)$. But on the other hand Label Smoothing leads to the extra terms $\frac{\sigma}{c-1}\log( \prod_{j, j\neq i} p ( y = y_j | \mathbf{r}))$. These terms encourage the most discriminant features for the other classes $j$. So instead of reinforcing only the discriminant features w.r.t the relevant classes, Label Smoothing will also encourage discriminant features of all classes. 

Hence Label Smoothing promotes diversity but only among discriminant features. Features leading to relevant accuracy  $( p ( y = y_i | \mathbf{r}) > 0.5)$ will not be encouraged. This is congruent with \cite{islabelsmoothingtrully,whenlabelsmoothinghelp} where it is reported that Label Smoothing erases intra-class information to promote inter-class information.

\subsection{Entropy regularization to encourage diversity in the feature space}
Supported by experiments showing the correlation between the entropy of the feature space and the regression ability we investigate the use of entropy of the features as a regularization term.
While Cross Entropy and Label Smoothing tend to support the most discriminant features we show that entropy regularization on the features promotes diversity. 

The idea is to directly add the entropy of the feature space as a regularization term. Within the Bayesian framework we can write our criterion:
\begin{align}
   \mathcal{L}_{Entropy}^{Bayesian}(\mathbf{x},\mathbf{y},\boldsymbol{\theta}) =  \mathbb{E}_{\mathbf{x}} \left[ - \mathbb{E}_{\mathbf{r} \sim p_{\boldsymbol{\theta}}(\mathbf{r} |\mathbf{x})} [ \log ( p ( y = y_i | \mathbf{r})) ] \right] - \lambda H_{\boldsymbol{\theta}}(\mathbf{r}).
\end{align}
where the entropy $H_{\boldsymbol{\theta}}(\mathbf{r})$ is estimated for each batch and $\lambda >0 $ is a scalar hyperparameter. When derivating the gradient of $H_{\boldsymbol{\theta}}(\mathbf{r})$, we obtain:
\begin{align*}
    \nabla_{\boldsymbol{\theta}}(- H_{\boldsymbol{\theta}}(\mathbf{r})) &= \nabla_{\boldsymbol{\theta}}  \int p_{\boldsymbol{\theta}}(\mathbf{r})  \log(p_{\boldsymbol{\theta}}(\mathbf{r})) d\mathbf{r} =  \nabla_{\boldsymbol{\theta}}  \int \mathbb{E}_{\mathbf{x}} \left [p_{\boldsymbol{\theta}} (\mathbf{r} |\mathbf{x}) \log(p_{\boldsymbol{\theta}}(\mathbf{r}))d\mathbf{r} \right] \\
    &=  \mathbb{E}_{\mathbf{x}} \left [ \int \nabla_{\boldsymbol{\theta}} (p_{\boldsymbol{\theta}} (\mathbf{r} |\mathbf{x} )) (\log p_{\boldsymbol{\theta}} (\mathbf{r}))] d\mathbf{r}\right]  + \int p_{\boldsymbol{\theta}}(\mathbf{r}) \frac{1}{p_{\boldsymbol{\theta}}(\mathbf{r})} \nabla_{\boldsymbol{\theta}} (p_{\boldsymbol{\theta}} (\mathbf{r})) d\mathbf{r} \\ 
    &=   \mathbb{E}_{\mathbf{x}} \left [ \int (\nabla_{\boldsymbol{\theta}}  (p_{\boldsymbol{\theta}} (\mathbf{r} |\mathbf{x})) (\log p_{\boldsymbol{\theta}} (\mathbf{r})) +  \nabla_{\boldsymbol{\theta}} p_{\boldsymbol{\theta}} (\mathbf{r} |\mathbf{x} )) d\mathbf{r} \right].
\end{align*}

So the final gradient of Cross Entropy with entropy regularization is:
\begin{align}
    -\mathbb{E}_{\mathbf{x}} \left[  \int \nabla_{\boldsymbol{\theta}} (p_{\boldsymbol{\theta}}( \mathbf{r} |\mathbf{x})) \left[ \log ( p ( y = y_i | \mathbf{r})) -\lambda ( \log (p_{\boldsymbol{\theta}} (\mathbf{r})) + 1)  \right]  d\mathbf{r} \right].
\end{align}
 
Notice that if a feature is very likely, i.e $p(\mathbf{r})$ large, the gradient will be penalized. The other term can be seen as a constant regularization. Tuning $\lambda$ will force the model to keep less discriminant features as long as they remain relevant for the coarse label classification task. 

\section{Implementation of features entropy regularization }
\subsection{Estimation of the features entropy}

With FIERCE, we propose to add the negative entropy of the features $\tilde H_{\boldsymbol{\theta}} (\mathbf{ \tilde r})$ to the Cross Entropy:
\begin{align*}
    \mathcal{L}_{Entropy}(\mathbf{x},\mathbf{y},\boldsymbol{\theta}) =  \mathcal{L}_{CE}(\mathbf{x},\mathbf{y},\boldsymbol{\theta}) - \tilde H_{\boldsymbol{\theta}}(  \mathbf{ \tilde r}),
\end{align*}
where $\tilde H_{\boldsymbol{\theta}}(\mathbf{ \tilde r})$ is an approximation of the real entropy, $H_{\boldsymbol{\theta}}( \mathbf{r})$.

Unfortunately, there is no straightforward way to compute and differentiate the entropy in the feature space, as $p( \mathbf{r})$ must be computed first. To this end, let us assume that we can approximate the conditional distribution of the feature space $p_{\boldsymbol{\theta}}( \mathbf{r} |\mathbf{x})$ as a categorical distribution  $p_{\boldsymbol{\theta}} (\mathbf{ \tilde r}|\mathbf{x} )$ where $\mathbf{ \tilde r}$ is an approximation of the features $\mathbf{r}$. We recall that a categorical distribution is defined by a vector of probabilities $\pmb{\pi}$ s.t: $p_{\boldsymbol{\theta}}( \mathbf{ \tilde r} = \mathbf{ \tilde r}_i | \mathbf{x}) = \pmb{\pi}_i $.

To define categories we use anchor points  $\mathbf{ \tilde r}$ from an embedding space $\mathcal{E} \in \mathbb{R}^{e \times d}$ where $e$ is number of anchor points  (vectors) and $d$ is the dimension of the feature space. Each feature can be mapped to the $i$-th entry $\mathbf{ \tilde r}_i \in \mathcal{E}$ sharing the highest similarity (cosine). We define the probability of these mappings, i.e the categorical probability $p_{\boldsymbol{\theta}} (\mathbf{ \tilde r} = \mathbf{ \tilde r}_i |\mathbf{x} )$, by the values of the following one hot tensor:
\begin{align}
  \label{eq:onehotcat}
    p_{\boldsymbol{\theta}}^{\text{one\_hot}}(\mathbf{ \tilde r} = \mathbf{ \tilde r}_i |\mathbf{x}) = \text{one\_hot}(\text{argmax}_j sim( \mathbf{r},\mathbf{ \tilde r}_j)) [i] =\left\{ 
                \begin{array}{ll}
                  1 \text{ if } i =  \text{argmax}_j sim( \mathbf{r}, \mathbf{\tilde r}_j),\\
                  0 \text{ otherwise.}\\
                \end{array}
              \right.
\end{align}
The probability of each discrete feature is then empirically estimated over each batch by $p_{\boldsymbol{\theta}}(\mathbf{ \tilde r} = \mathbf{ \tilde r}_i) = \frac{1}{N} \sum_x p( \mathbf{r} = i |\mathbf{x})$ where $N$ is the size of the batch. The anchor points are fixed and uniformly randomly sampled at the beginning of the training. A differentiable way to update these anchor points  to better align with the observed features is left as future work.


\subsection{Differentiation of the feature entropies}
Equation \ref{eq:onehotcat} above is not differentiable, meaning that the gradient from $p_{\boldsymbol{\theta}}(\mathbf{\tilde r})$ is not available. To overcome this difficulty we redefine $ p_{\boldsymbol{\theta}}(\mathbf{ \tilde r} = \mathbf{ \tilde r}_i |\mathbf{x})$. We use a soft-one-hot tensor sampled from the softmax-gumbel function~\cite{gumbel1,gumbel2} that allows auto-differentiation. This method relies on the Gumbel-Max trick~\cite{gumbel1} that provides an efficient way to sample from a categorical distribution: 
\begin{align}
    p_{\boldsymbol{\theta}} (\mathbf{ \tilde r}  = i |\mathbf{x} ) = \text{one\_hot}(\text{argmax}_j[ \log( \pmb{\pi}_i )+ g_i]), 
\end{align}
where $g_1,g_2, \cdots, g_e$ are i.i.d drawn from a Gumbel distribution: Gumbel$(0,1)$. To get an equation which is differentiable, we approximate as in~\cite{gumbel2} the $\text{argmax}$ with the $softmax$ function:
\begin{align}
\label{eq:softonehotcat}
    p_{\boldsymbol{\theta}}^{soft\_\text{one\_hot}}  (\mathbf{ \tilde r} = i |\mathbf{x} ) = \frac{exp((log( \pmb{\pi}_i )+ g_i)/\tau)}{\sum_j exp((log( \pmb{\pi}_i )+ g_i) / \tau)}
\end{align}
As the softmax temperature $\tau$ approaches 0, samples from the Gumbel-Softmax distribution become one-hot and the Gumbel-Softmax distribution becomes identical to the categorical distribution $p(\mathbf{z})$. Futhermore, one can get a one-hot tensor by using the following trick: 
\begin{align}
    p (\mathbf{ \tilde r} = \mathbf{ \tilde r}_i |\mathbf{x} )  = sg(p_{\boldsymbol{\theta}}^{\text{one\_hot}}(\mathbf{ \tilde r} = \mathbf{ \tilde r}_i |\mathbf{x} ) - p_{\boldsymbol{\theta}}^{\text{soft\_one\_hot}}(\mathbf{ \tilde r} = \tilde r_i |\mathbf{x} )) + p_{\boldsymbol{\theta}}^{\text{soft\_one\_hot}}(\mathbf{ \tilde r} = \mathbf{ \tilde r}_i | \mathbf{x} ).
\end{align}
where $sg$ is the stop gradient operator, $p_{\boldsymbol{\theta}}^{\text{one\_hot}}$ is given by Eq~\ref{eq:onehotcat} and $p_{\boldsymbol{\theta}}^{\text{soft\_one\_hot}}$  by Eq.~\ref{eq:softonehotcat}. The stop gradient operator is a standard operator of automatic differentiation libraries, such as Tensorflow or PyTorch. It produces an expression with no gradient but a normal value on the forward pass.

In our case the component of the vector $\pmb{\pi}$ contains the similarity between the features $\mathbf{r}$ and each entry $\mathbf{\tilde r}_i$ of the embedding space.


\section{Experiments}
\label{experiment}
In this part we show the benefits of the FIERCE method on multiple datasets and benchmarks. We first show that the entropy of the features is correlated to the ability of the feature space to be recycled and to unveil fine labels. We investigate the use of the output distribution as an indicator of the transfer ability of the feature space.
Eventually, we explore the use of FIERCE in transfer learning on few-shot datasets. We carry out all the experiments on a 3090 GPU and estimate the amount of total computation time needed to reproduce the experiments to be about $500$ hours.
\subsection{Correlation between the entropy of the feature space and transfer ability}
Considering a regression problem turned into a classification one, we experimentally show that the entropy of the feature space is correlated with the regression ability (MSE) in the feature space.

Let us consider a regression problem with a dataset $\mathcal{D}^{reg}$. We transform $\mathcal{D}^{reg}$ into a coarse classification problem where only coarse labels are available. The dataset is split intro train and test: $\mathcal{D}_{train}^{classification} = \{(\mathbf{x}_i,\mathbf{y}_i = coarsen(\mathbf{z}_i)), \ 1 \leq i \leq N_{train}\} $, where $coarsen$ is a function that turns its fine input into a coarse output. 

Our purpose is to train a model using the coarse labels from $\mathcal{D}_{train}^{classification}$ and then recycle the feature space to infer the fine labels of  $\mathcal{D}^{reg}$. We compare 3 criteria: Cross Entropy, Label Smoothing, and our proposed method Feature Information Entropy Regularized Cross Entropy, FIERCE. Note that we did not compare with Confident penalization as it is  similar to Label Smoothing.

We take the dataset of age estimation~\cite{ageestimation} based on face photos. The goal is to predict the ages corresponding to the face images.  The dataset can be turned into a binary classification problem with the coarse labels $\mathbf{y} = \text{one\_hot}( \mathds{1}_{age<36})$ where 36 is the median age. Note that this regression dataset has already been turned into a classification dataset by splitting the ages into much more classes (\cite{ageintoregression,ageintoregressionandclassif}). We use a Resnet-18~\cite{resnet}, except that at the penultimate layer we average the features maps to get a feature space of dimension 1. We expect that a 1 dimensional feature space preserves the order relation that exists for the true age labels to some extent.

To estimate the fine-grained ages we compute the features of 10000 samples. We rank the features by their values and, assuming that the probability distribution of the ages is known, we map it to the distribution of the features using an Optimal Transport 1d mapping~\cite{optimalvillani}. For more robust results, we interpolate this prediction with the average age of individuals in the considered class. For each experiment we take the coefficient interpolation which provide the lowest MSE. An interpolation coefficient of 1 means that no information from the feature space is used: ages are predicted directly from the class average age. This latter computation acts as a natural baseline as it is the best that can be achieved without access to the face images.

We report the evolution of the MSE and the entropy on Figure \ref{fig:EvolutionEntropyMSE}. We use Stochastic Gradient Descent (SGD). When using Label Smoothing or raw Cross-Entropy, we observe two phases. First the MSE decreases while the accuracy improves. Then a much longer second phase starts (around epoch 50)  where the MSE increases while the accuracy remains stable. Similarly, the entropy of the feature space reaches a maximum before decreasing, and the decrease is negatively correlated with the drop in regression ability. These phases are very similar to evolution of the mutual information $I(\mathbf{r},\mathbf{x})$  described in \cite{blackbox}, and a strong motivation for the introduction of FIERCE.

We observe that our method provides the lowest MSE and impacts the entropy of the feature space as the entropy increases around epoch 200. In the appendix we show the evolution of the MSE and the interpolation coefficient with respect to the hyperparameter $\lambda$. One can see that tuning $\lambda$ allows to retrieve more and more information on the coarse labels from the feature spaces as the optimal coefficient of interpolation keeps increasing, acting as more evidence of the relationship between entropy in the latent space and the ability to generalize to finer labels.

We note on this dataset that Label Smoothing does not perform well: the final MSE is worse than the one given by Cross Entropy and the entropy keeps dropping. In the appendix we show the feature space of each criteria. One can clearly see how Label Smoothing is increasing the uncertainty area  without providing relevant information on the coarse labels.

\subsection{Output distribution and transferability}
Entropy regularization techniques on the output distribution reduces the peaks and smooths the distribution \cite{penalisingconfident,labelsmoothing}. Instead of having outputs either close to 0 and 1, smoothing the distribution allows to have a larger spectrum of output values. A natural question is then: does this smoothness provide information on fine labels?

Previous authors interpreted the output values as the uncertainty of the predictions \cite{calibration}. An other interpretation is to consider the output values as information on the fine labels; for instance it could be the degrees of similarity with each class.

To investigate the validity of this interpretation, we consider a dataset where the output uncertainty can be easily interpreted as the fine labels. For that matter, we take a dataset made of crop from a single hyperspectral image (remote sensing) ~\cite{ghamisi2017advances}. We detail the dataset in appendix. It is composed of $w \times h$ crop from a hyperspectral image gathering reflectance values for $c$ contiguous wavelengths (channels) in the visible and near infrared domains. Two important interconnected problems in hyperspectral imaging are supervised semantic pixel classification and spectral unmixing~\cite{ghamisi2017advances}. In classification, one tries to assign a class to every pixel among a predefined number of identified classes. Unmixing can be seen as a refinement of classification where one accounts for the fact that objects to be detected may be smaller than the size of a pixel. Then the goal is to predict the proportion of each material (called abundances) in each pixel. We give more details on this dataset in appendix.

We turn the regression problem of unmixing into a classification problem by using the 
coarse labels $\mathbf{y} = \text{one\_hot}(\text{argmax}_i \ \mathbf{z}_i)$ where $\mathbf{z}_i $ is the proportion of material $i$:   $\mathbf{z} \in [0,1]^{m}, \sum_i (\mathbf{z}_i) = 1 $.  For this dataset, we use a fully connected neural network $f_{\boldsymbol{\theta}}$ with 2 hidden layers where $\boldsymbol{\theta}$ are the parameters of the network. 
As $\mathbf{y}$ are proportions, one can interpret the class probabilities of the network, ie $(q_{\boldsymbol{\theta}}( \mathbf{y}|\mathbf{x} )$, as the proportions of each material. We can evaluate the regression performance of the DNN with the mean square error loss over the whole image:
\begin{align}
     raw \ MSE(f_{\boldsymbol{\theta}}) = \mathbb{E}_{\mathcal{D}^{reg}} [\sum_i q_{\boldsymbol{\theta}}( \mathbf{y}|\mathbf{x} )_i - \mathbf{z}_i)^2]^{1/2}.
\end{align}

We compute another metric to evaluate whether the inferred feature spaces can be recycled to perform the regression task. For each criterion, we retrain the network with a regression criterion (Mean Square Error). We use and keep fixed the feature space inferred with the classification criteria and we compute what we call the transfer Mean Square Error: $transfer \ MSE(f_{\boldsymbol{\theta}})$.

As reported in Table~\ref{tab:stats}, Label Smoothing and FIERCE demonstrate the strongest ability to recover the regression labels directly from the output with the lowest MSE (averaged over all materials). But, we note that FIERCE presents the highest transfer ability (lowest transfer MSE) meaning that its feature space allows to retrieve more information on the fine labels than other criteria. Considering the difference between the two metrics --raw MSE and transfer MSE-- we suspect that the output distribution is not an accurate indicator of the ability in transfer.

\begin{table}
    \caption{MSEs estimated on the hyperspectral dataset for the different criteria. Raw MSE given directly from the output of the network, transfer MSE given by transfer learning.}
    \label{tab:stats}
    \centering
    \begin{tabular}{llll}
    \toprule
        & Cross Entropy  & Label Smoothing ($\sigma = 0.4$) &  FIERCE ($\lambda = 1$) \\ 
       \midrule
        Raw MSE  & $0.186 \pm 0.001$ & {$\mathbf{0.066}\pm 0.1$} & $0.130 \pm 0.008$ \\
        Transfer MSE & $0.177 \pm 0.3$ & $0.017 \pm 0.08$ & $\mathbf{0.006} \pm 0.001$ \\
        \bottomrule
    \end{tabular}
\end{table}

We then consider multiple metrics defined on the output distribution, as a mean to better distinguish Label Smoothing from our method. We use: reliability diagrams, \emph{Expected Calibration Error} (ECE), Maximum Calibration Error (MCE)~\cite{mce}, the mutual information between the input and the ouput as suggested in ~\cite{whenlabelsmoothinghelp}, and another metric called Stability which indicates how much a prediction within a class can fluctuate. We report the corresponding results in appendix.

We observe that Label Smoothing and FIERCE present similar values of mutual information and Stability. The mutual information of these two criteria is smaller that the mutual information given by the Cross Entropy. For Label Smoothing, the result was expected as this method erases the information contained in the different similarities that an individual share with the different classes ~\cite{whenlabelsmoothinghelp,islabelsmoothingtrully} resulting in lowest mutual information.
But the result is surprising for our method as it presents the smallest transfer MSE. Hence, it seems that using the output distribution as an indicator of the generality of the feature space is not satisfying as it is not sufficient to discriminate our method from Label Smoothing. In the appendix we visualize the different feature spaces and show that FIERCE could allow to retrieve more information on the semantic similarity. 




A conclusion is that the output distribution is not an accurate indicator of the information contained in the feature space: the raw MSE seems to indicate that FIERCE and Label Smoothing would perform similarly in transfer. The transfer MSE of Cross-entropy is really high and worse than the raw MSE which could indicate that the feature space is overspecialized and the feature space too biased.
\subsection{Transfer learning: fewshot applications}
As the previous experiments demonstrated, the FIERCE method allows to retrieve more information from the feature space than Cross Entropy or Label Smoothing. This is is not necessarily relevant in classification where one wants to select only the features meaningful to discriminate the different classes. But we think that the proposed method could be beneficial in transfer learning where the feature space can be reused to perform different, possibly more subtle, tasks. In the following, we present results when testing FIERCE on different fewshot problems \cite{fewshot1}. In fewshot learning, a model is first trained on a generic dataset. Then the goal is to use the feature space to classify new categories with a limited number of samples.

 \begin{table}
      \caption{Classification accuracy (1-shot) on novel tasks for CIFAR-FS and CUB. Displayed accuracy are averaged over 10,000 random fewshot runs. Confidence intervals are computed over 10 randomly initialized training of each model under each criterion.}
     \label{tab:my_label}
     \centering
     \begin{tabular}{llll}
     \toprule
          & Cross Entropy & Label Smoothing & FIERCE  \\
        \midrule
        CIFAR-FS & $64.32 \pm 0.7  $  & $65.76 \pm 0.57$ & $\mathbf{66.16}\pm 1.04$  \\
        CUB & $59.61 \pm 0.84$ &$59.87\pm 0.81$  &$\mathbf{62.18} \pm 0.59$ \\
    \bottomrule
     \end{tabular}
 \end{table}

We evaluate our method on two datasets:  on CIFAR-FS\cite{cifarfs} with a Resnet18 and on CUB~\cite{cub}. We compare the 3 criteria: Cross Entropy, Label Smoothing and FIERCE.
CIFAR-FS is a dataset made of 60,000 color (RGB) images of sizes 32x32 pixels. The dataset is divided in 3 splits of 64 training classes, 16 validation classes, and 20 testing classes each containing 600 examples. 
CUB is composed of 11,788 images of size 84×84 pixels and 200 classes. We use the splits: 100 bases classes, 50 novels recommended in ~\cite{splitCUB}.
On both datasets we outperform the baseline and Label Smoothing. Interestingly, while Label Smoothing and FIERCE have very different impact on the feature space they both improve the accuracy in transfer.
\section{Limitations}
\label{limitations}
We observed that the entropy of the features space was related to the transfer ability, especially in the case of regression. But, the entropy may be not the optimal criterion to use as a regularizer.

On regression datasets, our method yield the lowest MSEs yet they do not significantly outperform those achieved by raw Cross Entropy during the first epochs of the training. The main interest of the proposed method is thus the added ability to avoid early stopping.

On another note, one difficulty present in our method is the computation of a high dimension feature space. This computation must also be auto-differentiable. Our estimation relies on random anchor points and the generation and distribution of these anchor points may have an impact on the geometry of the feature space: with high values of $\lambda$ we force the features to be uniformly aligned with these anchor points. One solution would be the use of learnable dictionary such as it is performed with VQ-VAE~\cite{VQVAE} but that would also mean adding an extra loss and another parameter.

Finally, let us add that the hyperparameter $\lambda$ may be hard to tune in practice, especially in domains where there is no clear insights about future use of the coarsely trained model.

\section{Conclusion}
In this paper, we introduced a new regularization term which applies on the feature space of a deep learning architecture. Its aim is to provide a richer feature space that can be reused later on to solve other tasks. To this end, it promotes entropy among the features, preventing the model from removing too many of them, even though their help in solving the training task is moderate.

We proposed a trick to evaluate the entropy in the feature space, relying on an embedding through a fixed number of random anchor points, and the use of a softmax-gumbel function.
Supported by empirical validation and theoretical development, we demonstrated the ability of the proposed method to prevent the loss of information caused by using raw cross-entropy or even label smoothing, leading to better performance on fine-grain problems.


\bibliographystyle{plainnat}
\bibliography{main}
\section*{Checklist}

\begin{enumerate}

\item For all authors...
\begin{enumerate}
  \item Do the main claims made in the abstract and introduction accurately reflect the paper's contributions and scope?
    \answerYes{}
  \item Did you describe the limitations of your work?
    \answerYes{ See Section~\ref{limitations}}
  \item Did you discuss any potential negative societal impacts of your work?
    \answerNo{}
  \item Have you read the ethics review guidelines and ensured that your paper conforms to them?
    \answerYes{}
\end{enumerate}

\item If you are including theoretical results...
\begin{enumerate}
  \item Did you state the full set of assumptions of all theoretical results?
    \answerYes{}
        \item Did you include complete proofs of all theoretical results?
    \answerYes{}
\end{enumerate}

\item If you ran experiments...
\begin{enumerate}
  \item Did you include the code, data, and instructions needed to reproduce the main experimental results (either in the supplemental material or as a URL)?
    \answerYes{as supplementary materials  }
  \item Did you specify all the training details (e.g., data splits, hyperparameters, how they were chosen)?
    \answerYes{we provide the ablation studies on the hyperparameter in appendix}
        \item Did you report error bars (e.g., with respect to the random seed after running experiments multiple times)?
    \answerYes{we report the mean averaged over 10 runs and the confidence intervals}
        \item Did you include the total amount of compute and the type of resources used (e.g., type of GPUs, internal cluster, or cloud provider)?
    \answerYes{See Section ~\ref{experiment}}
\end{enumerate}

\item If you are using existing assets (e.g., code, data, models) or curating/releasing new assets...
\begin{enumerate}
  \item If your work uses existing assets, did you cite the creators?
    \answerYes{}{}
  \item Did you mention the license of the assets?
    \answerYes{}{Licence are given by the references}
  \item Did you include any new assets either in the supplemental material or as a URL?
    \answerYes{}
  \item Did you discuss whether and how consent was obtained from people whose data you're using/curating?
    \answerNA{}
  \item Did you discuss whether the data you are using/curating contains personally identifiable information or offensive content?
    \answerNA{}
\end{enumerate}

\item If you used crowdsourcing or conducted research with human subjects...
\begin{enumerate}
  \item Did you include the full text of instructions given to participants and screenshots, if applicable?
    \answerNA{}
  \item Did you describe any potential participant risks, with links to Institutional Review Board (IRB) approvals, if applicable?
    \answerNA{}
  \item Did you include the estimated hourly wage paid to participants and the total amount spent on participant compensation?
    \answerNA{}
\end{enumerate}

\end{enumerate}


\end{document}


\appendix

\section{Appendix}
Licence, Code and Datasets used for the Experiments and Appendix can be found in the following repository: \url{https://anonymous.4open.science/r/FIERCE-repo-CFE2/README.md}.
\subsection{Sensitivity analysis}
\begin{figure}[htpb!]
    \centering
    \subfigure{
    \begin{tikzpicture}
      \pgfplotsset{%
    width=.1pt,
    height=.2\textwidth
}
        \begin{axis}[
        width=6cm,
        height = 4cm,
            xlabel={$\lambda$},
            ylabel={MSE},
            xmode=log,
            ymode= log,
            xmin=0.01, xmax=10,
            ymin=0.001, ymax=0.2,
            xtick={0.01,0.1,0.1,1,10},
            ytick={0.001,0.01,0.1},
            legend pos=north west,
            ymajorgrids=true,
            grid style=dashed,
        ]
        
        \addplot[
            color=blue,
            mark=square,
            ]
            coordinates {
            (0.01,0.02)(0.1,0.0036)(1,0.006)(2,0.0065)(3,0.0074)(4,0.009)(6,0.0089)(8,0.012)(10,0.016)
            };

        \addplot[
            color=red,
            mark=square,
            ]
            coordinates {
            (0.01,0.18)(0.1,0.168)(1,0.1292)(2,0.1036)(3,0.0876)(4,0.08633)(6,0.1335)(8,0.1854)(10,0.256)
            };

        \end{axis}
        \end{tikzpicture}
        
        }
        \subfigure{
        
        \begin{tikzpicture}
          \pgfplotsset{%
    width=.1pt,
    height=.2\textwidth
}
    
        \begin{axis}[
        width=6cm,
        height = 4cm,
            xlabel={$\lambda$},
            xmode=log,
            xmin=0.01, xmax=1,
            ymin=53, ymax=66,
            xtick={0.01,0.1,1},
            ytick={55,60,65},
            legend pos=north west,
            ymajorgrids=true,
            grid style=dashed,
        ]
        
        \addplot[
            color=red,
            mark=square,
            ]
            coordinates {
            (0.01,63.66)(0.03,65.27048)(0.05,61.58)(0.1,61.07806)(0.2,56.53666)(0.3,55.74377)(0.5,66)
            };

        \end{axis}
        \end{tikzpicture}

        }
   
    \caption{Evolution of the MSE with respect to $\lambda$. On the left the hyperspectral dataset: raw MSE (red), transfer MSE (blue). On the right the age estimation dataset.}
    \label{fig:evolutionMSelambda}
\end{figure}
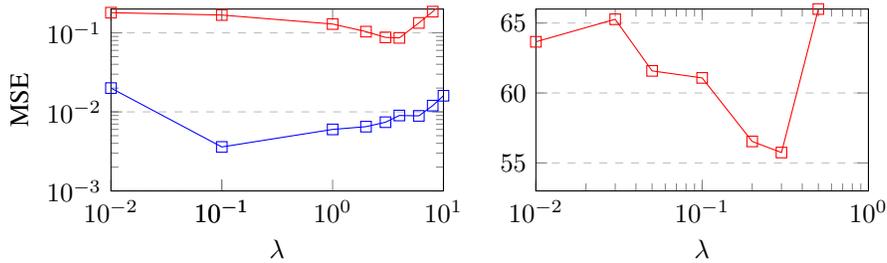

In this section we study the effect of the choice of the hyperparameters in the experiments reported in Section 5. Let us recall that FIERCE introduces two hyperparameters: $\lambda$, tuning the influence of the entropy constraint and $e$, the number of anchor points of $\mathcal{E}$ used to approximate the entropy of the features.
\paragraph{Influence of $\lambda$:}
On Figure~\ref{fig:evolutionMSelambda} we plot for each dataset the influence of $\lambda$. For the hyperspectral dataset we report the evolution of the raw MSE and the transfer MSE; for age estimation the evolution of the MSE. We observe that the best choice of $\lambda$ is not consistent for the raw MSE and transfer MSE. Yet, in this experiment, a fairly large range of values yield interesting results. In the case of the age estimation dataset, we note that the best choice for $\lambda$ is the limit beyond which the classification score dramatically reduces. In both experiments, we point out the existence of an optimal nondegenerate value of $\lambda$, motivating for its importance.

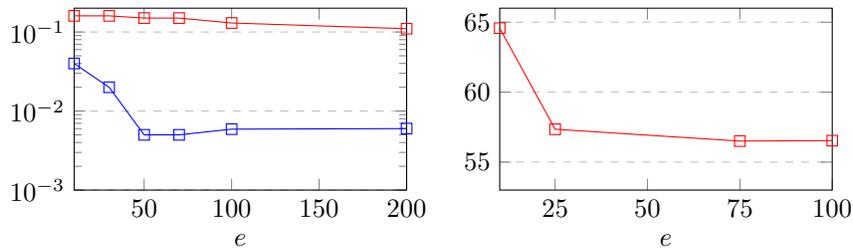
\begin{figure}[!htpb]
    \centering
    \subfigure{
    \begin{tikzpicture}
      \pgfplotsset{%
    width=.1pt,
    height=.2\textwidth
}
        \begin{axis}[
        width=6cm,
        height = 4cm,
            xlabel={$e$},
            ymode= log,
            xmin=10, xmax=200,
            ymin=0.001, ymax=0.2,
            xtick={50,100,150,200},
            ytick={0.001,0.01,0.1},
            legend pos=north west,
            ymajorgrids=true,
            grid style=dashed,
        ]
        
        \addplot[
            color=red,
            mark=square,
            ]
            coordinates {
            (10,0.16)(30,0.16)(50,0.15)(70,0.15)(100,0.13)(200,0.11)
            };

        \addplot[
            color=blue,
            mark=square,
            ]
            coordinates {
            (10,0.040)(30,0.02)(50,0.005)(70,0.005)(100,0.0059)(200,0.006)
            };

        \end{axis}
        \end{tikzpicture}
        
        }
        \subfigure{
        
        \begin{tikzpicture}
          \pgfplotsset{%
    width=.1pt,
    height=.2\textwidth
}
    
        \begin{axis}[
        width=6cm,
        height = 4cm,
            xlabel={$e$},
            xmin=10, xmax=100,
            ymin=53, ymax=66,
            xtick={0,25,50,75,100},
            ytick={50,55,60,65,70},
            legend pos=north west,
            ymajorgrids=true,
            grid style=dashed,
        ]
        
        \addplot[
            color=red,
            mark=square,
            ]
            coordinates {
            (10,64.57625)(25,57.34856)(75,56.50416)(100,56.53666)
            };

        \end{axis}
        \end{tikzpicture}
        
        }
   
    \caption{Evolution of the MSE with respect to the number of entries $e$. On the left the hyperspectral dataset: raw MSE (red), transfer MSE (blue). On the right age estimation. Minimum is $e= 10$.}
    \label{fig:evolutionwithembedding}
\end{figure}

\paragraph{Influence of the number of anchor points $e$:}
On Figure~\ref{fig:evolutionwithembedding} we plot for the hyperspectral and age estimation datasets the influence of $e$ on the MSEs.
We note that after a certain limit, $e = 100$ for the hyperspectral dataset and $e = 75$ for age estimation, the MSE remains stable. These values are around the batch size for each dataset: $200$ for the hyperspectral dataset and $64 $ for age estimation. Therefore, we take for the other experiments a value of $e$ close to the batch sizes.

\subsection{Metrics to evaluate the output distribution}
On the hyperspectral and age estimation datasets, we compute metrics introduced by previous authors to evaluate the output distribution: reliability diagram, stability, mutual information. We report the metrics in Table ~\ref{table:outputstats}.
 On the hyperspectral dataset, FIERCE presents the lowest mutual information while achieving a raw MSE lower than the Cross Entropy and the lowest transfer MSE. On the age estimation data, FIERCE presents the highest mutual information. To some extent, Mutual Information and Stability seem related but they both failed to distinguish the criteria: on the hyperspectral dataset, Label Smoothing and FIERCE have a very similar Stability. On the age estimation data, all metrics are similar for each criterion, while the MSEs are quite different (See Section 5.2). A conclusion is that these metrics do not seem reliable when it comes to evaluate the ability in transfer.

\begin{figure}[!h]
\centering
\subfigure[Hyperspectral: Cross Entropy]{
     \includegraphics[width=0.3\textwidth]{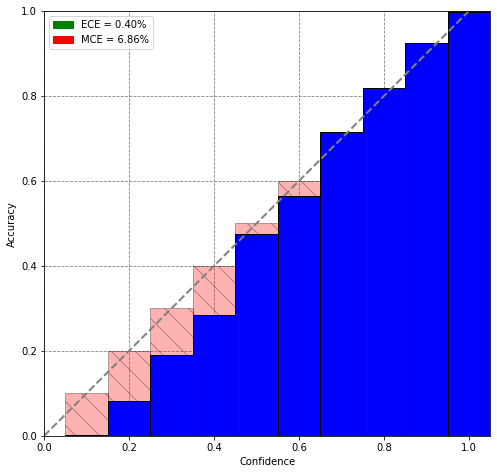}
    \label{figscc:subfig1}
}
\subfigure[Hyperspectral: Label Smooth.]{
    \includegraphics[width=0.3\textwidth]{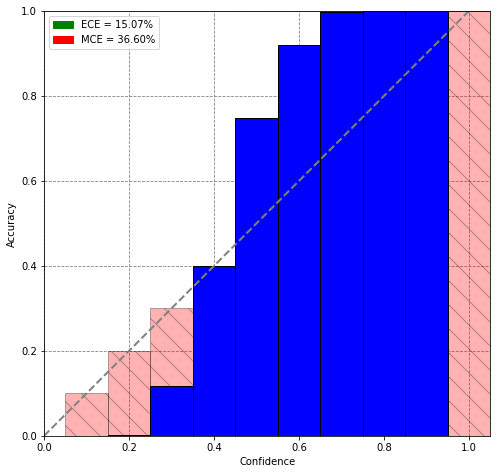}
    \label{figscc:subfig2}
}
\subfigure[Hyperspectral: FIERCE]{
    \includegraphics[width=0.3\textwidth]{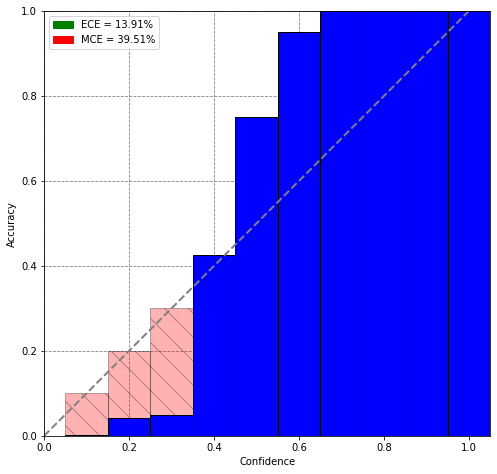}
    \label{figscc:subfig3}
}
\hfill
\subfigure[Age estimation: Cross Entropy]{
    \includegraphics[width=0.3\textwidth]{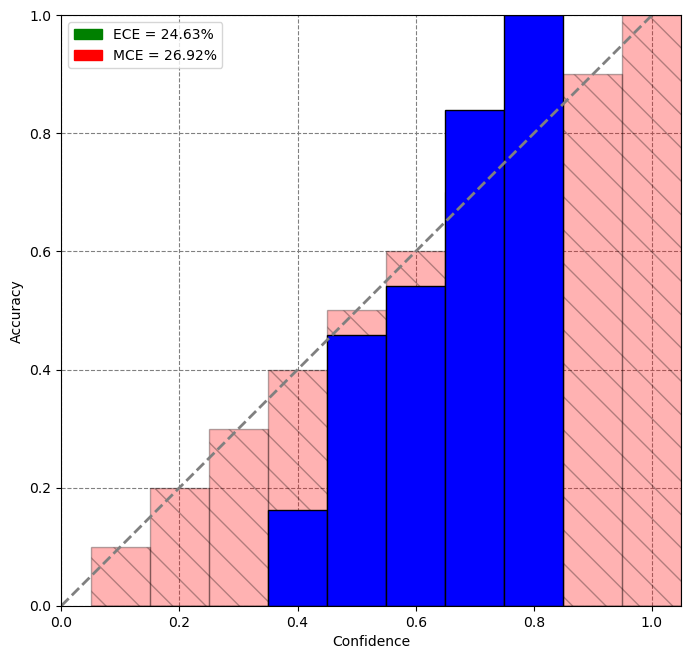}
    \label{figscc:subfig2}}
\subfigure[Age estimation: Label Smooth.]{
    \includegraphics[width=0.3\textwidth]{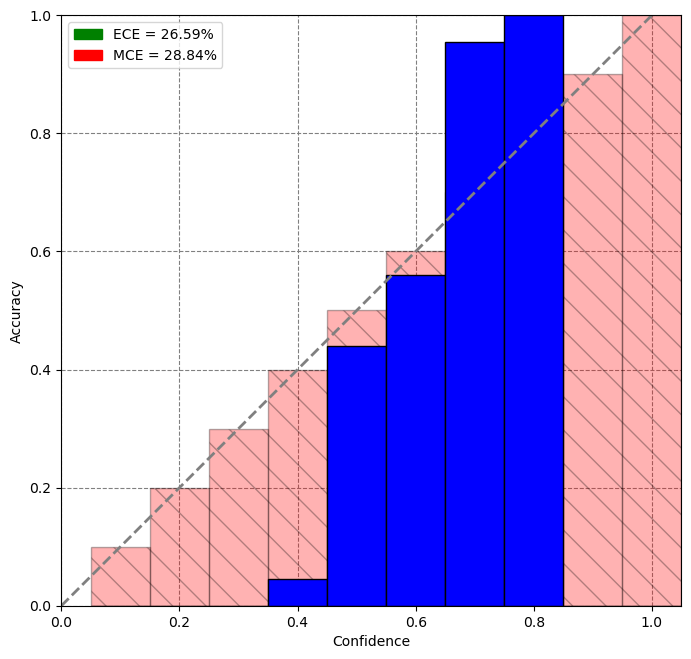}
    \label{figscc:subfig2}
}
\subfigure[Age estimation: FIERCE]{
    \includegraphics[width=0.3\textwidth]{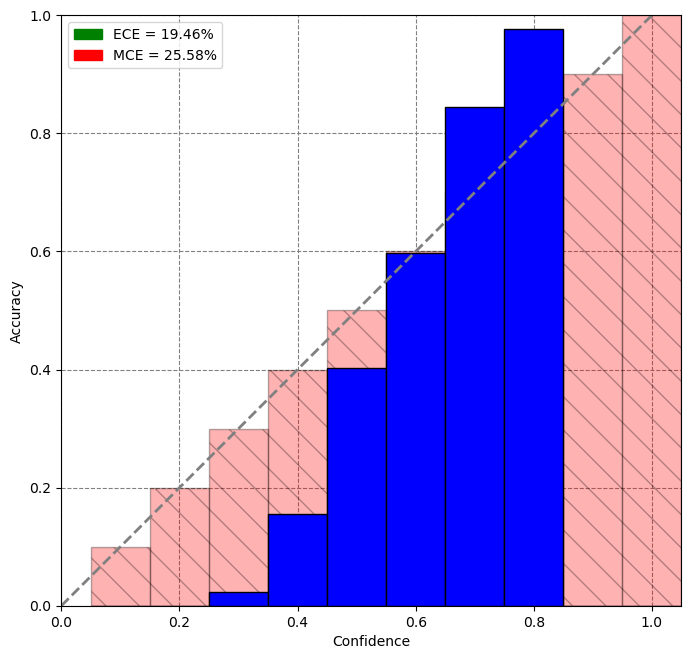}
    \label{figscc:subfig2}
}
\caption[Optional caption for list of figures]{Reliability diagrams for each criteria on the hyperspectral and age estimation datasets. }
\label{fig:reliabity}
\end{figure}

We show the reliability diagrams in Figure ~\ref{fig:reliabity}. We observe that neither Label Smoothing or FIERCE improve reliability. Label Smoothing seems to prevent confident prediction (output close to 1) which could explain why the raw MSE is smaller on the hyperspectral dataset. 

\begin{table}[!htpb]
    \caption{MSEs estimated on the hyperspectral dataset for the different criteria. Raw MSE given directly from the output of the network, transfer MSE given by transfer learning.}
    \label{tab:stats}
    \centering
    \begin{tabular}{lllll}
    \toprule
        & & Cross Entropy  & Label Smoothing &  FIERCE  \\ 
       \midrule
        Hyperspectral & Mutual Information  & $1.71$ & $0.55$ & $0.39$ \\
        & Stability & $0.88$ & $0.93$ & $0.93$ \\
        \midrule
        Age estimation & Mutual Information  & $0.10$ & $0.12$ & $0.14$ \\
        & Stability & $0.81$ & $0.80$ & $0.78$ \\
    \end{tabular}
    \label{table:outputstats}
\end{table}
\subsection{Influence of the criteria on the feature space}

\begin{figure}[!h]
\centering
\subfigure[Hyperspectral: Cross Entropy]{
     \includegraphics[width=0.3\textwidth]{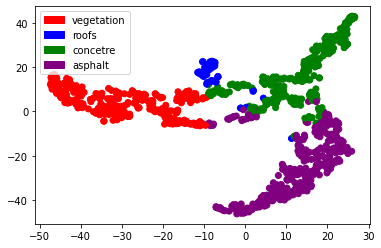}
    \label{figscc:subfig1}
}
\subfigure[Hyperspectral: Label Smooth.]{
    \includegraphics[width=0.3\textwidth]{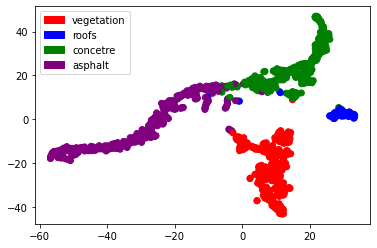}
    \label{figscc:subfig2}
}
\subfigure[Hyperspectral: FIERCE]{
    \includegraphics[width=0.3\textwidth]{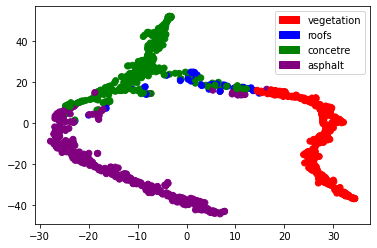}
    \label{figscc:subfig3}
}
\hfill
\subfigure[CIFAR-10: Cross Entropy]{
    \includegraphics[width=0.3\textwidth]{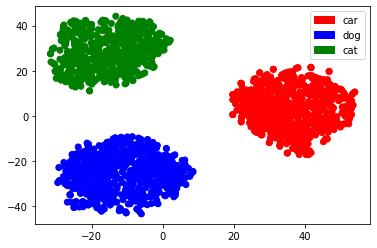}
    \label{figscc:subfig2}}
\subfigure[CIFAR-10: Label Smooth.]{
    \includegraphics[width=0.3\textwidth]{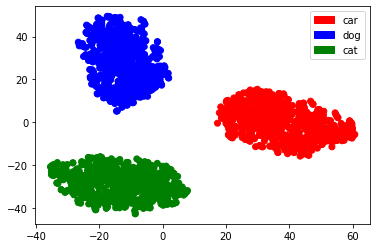}
    \label{figscc:subfig2}
}
\subfigure[CIFAR-10: FIERCE]{
    \includegraphics[width=0.3\textwidth]{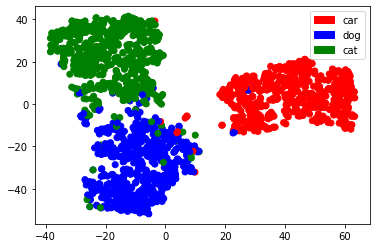}
    \label{figscc:subfig2}
}
\caption[Optional caption for list of figures]{TSNE representation of the feature space (penultimate layer) for the different criteria. First row on the hyperspectral dataset (dim = 100), second row on CIFAR-10.}
\label{figcc:representation}
\end{figure}

\paragraph{TSNE representation on CIFAR-10 and the hyperspectral dataset:} As an attempt to better understand the geometric impact of our proposed method, we embed features into a 2D space using t-Distributed Stochastic Neighbor Embedding (TSNE) \url{https://scikit-learn.org/stable/modules/generated/sklearn.manifold.TSNE.html}. We plot on Figure~\ref{figcc:representation} the different feature spaces that we obtained for each criterion on the hyperspectral dataset, and on CIFAR-10~ trained with a ResNet18. On the hyperspectral data, we can see that both the Cross Entropy and Label Smoothing split the feature representations of the different classes into different well separated clusters, while FIERCE preserves the continuity of the features from one class to the others. Indeed, transitions between clusters should correspond to mixed pixels, comprising a proportion of several materials (asphalt, concrete). On CIFAR-10, one can see that FIERCE reduces the distances between the different classes while smoothing the representation. Especially we observe than the intra-class distance between dogs and cats has been reduced compared to Label Smoothing or Cross Entropy. This reduction could indicate that the network is better at detecting a semantic similarities between classes. 

\begin{figure}[!h]
\centering
\subfigure[Cross Entropy]{
     \includegraphics[width=0.3\textwidth]{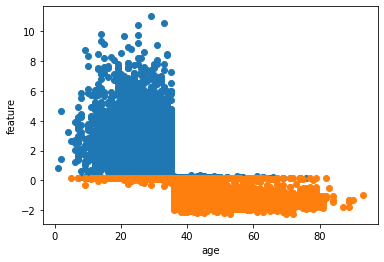}
    \label{figscc:subfig1}
}
\subfigure[Label Smoothing]{
    \includegraphics[width=0.3\textwidth]{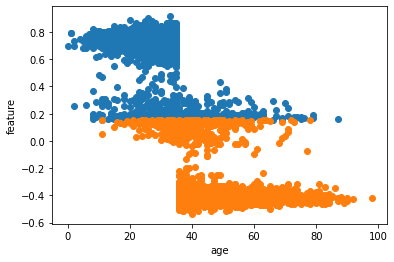}
    \label{figscc:subfig2}
}
\subfigure[FIERCE]{
    \includegraphics[width=0.3\textwidth]{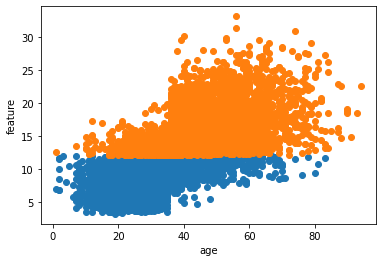}
    \label{figscc:subfig3}
}
\caption{Feature space ($1$-d) with respect to the ages for each criterion on the Age Estimation dataset.}
\label{fig:featuresAge}
\end{figure}

\paragraph{Age estimation:}
On the age estimation dataset, the dimension of the feature space is in $1$-d so we can directly represent the feature spaces in Figure ~\ref{fig:featuresAge}. Ideally the features should be aligned on a line or at least contained in an ellipse-like shape. 
For Cross Entropy, we observe that the features are well separated by the median of the ages (36) but randomly distributed in each part of the graph. For label smoothing, we observe that the area of uncertainty (in the middle) has increased but without improving the MSE. A conclusion is that this uncertainty does not provide relevant information on the ages: features are randomly distributed in this area. For FIERCE we observe than the features are much more aligned and are located in an ellipse, which is consistent with the lowest MSE that it achieves.
\subsection{Does our method approximate correctly the entropy?}
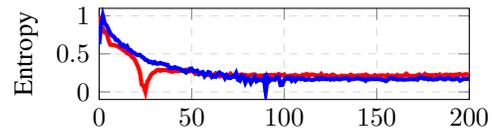
\begin{figure}[!htpb]
    \centering
      \begin{tikzpicture}
  \pgfplotsset{%
    width=.1pt,
    height=.2\textwidth
}
      \begin{axis}[
          width=6.5cm, 
          grid=major, 
          grid style={dashed,gray!30}, 
          ylabel= Entropy,
          legend style={at={(0.5,-0.2)},anchor=north}, 
        xmax=200,
        xmin=0,
        ]
        \addplot [  red,   ultra thick] table[x=Epoch,y=Approximation,col sep=comma] {compare_entropy.csv}; 
        \addplot [ blue,   ultra thick] table[x=Epoch,y=Entropy,col sep=comma] {compare_entropy.csv}; 

      \end{axis}
    \end{tikzpicture}
    \caption{Entropy of the feature space (in blue) and approximation of our method (in red) on the hyperspectral dataset.}
    \label{fig:Entropy}
\end{figure}

We investigate the quality of our entropy approximation. We carry out an experiment on the hyperspectral estimation with the Cross Entropy: we track the the Entropy of a feature space during the training and Entropy estimated by our method using the one hot tensor (not differentiable). We show the two measures on Figure ~\ref{fig:Entropy}. Entropy and Approximation are rescaled between 0 and 1.\\

Except for few anomalies, we observe that our approximation follows correctly the Entropy of the feature spaces. The correlation coefficient between the two measures is 0.86.